\documentclass[]{spie}  

\usepackage{amsmath,amsfonts,amssymb}
\usepackage{graphicx}
\usepackage{times}
\usepackage{epsfig}
\usepackage{paralist}
\usepackage{color}
\usepackage[colorlinks=true, allcolors=blue]{hyperref}
\usepackage{subfig}
\usepackage{dsfont}

\title{Augmentation techniques for video surveillance in the visible and thermal spectral range}

{\tt\small}

\author{Vanessa Buhrmester}
\author{Ann-Kristin Grosselfinger}
\author{David M\"unch}
\author{Michael Arens}
\affil{Fraunhofer IOSB, Gutleuthausstra{\ss}e 1, 76275 Ettlingen, Germany}

\authorinfo{Further information of correspondence author:\\
V.B.: E-mail: vanessa.buhrmester@iosb.fraunhofer.de, Telephone: +49 (0)72 43 92 21 67,\\
}

\begin{document} 
\maketitle

\begin{abstract}
In intelligent video surveillance, cameras record image sequences during day and night. Commonly, this demands different sensors. To achieve a better performance it is not unusual to combine them. We focus on the case that a long-wave infrared camera records continuously and in addition to this, another camera records in the visible spectral range during daytime and an intelligent algorithm supervises the picked up imagery. More accurate, our task is multispectral CNN-based object detection. At first glance, images originating from the visible spectral range differ between thermal infrared ones in the presence of color and distinct texture information on the one hand and in not containing information about thermal radiation that emits from objects on the other hand. Although color can provide valuable information for classification tasks, effects such as varying illumination and specialties of different sensors still represent significant problems. 
Anyway, obtaining sufficient and practical thermal infrared datasets for training a deep neural network poses still a challenge. That is the reason why training with the help of data from the visible spectral range could be advantageous, particularly if the data, which has to be evaluated contains both visible and infrared data. However, there is no clear evidence of how strongly variations in thermal radiation, shape, or color information influence classification accuracy. To gain deeper insight into how Convolutional Neural Networks make decisions and what they learn from different sensor input data, we investigate the suitability and robustness of different augmentation techniques. We use the publicly available large-scale multispectral ThermalWorld dataset consisting of images in the long-wave infrared and visible spectral range showing persons, vehicles, buildings, and pets and train for image classification a Convolutional Neural Network.
The training data will be augmented with several modifications based on their different properties to find out which ones cause which impact and lead to the best classification performance.

\end{abstract}


\section{INTRODUCTION}

\subsection{Motivation}
\label{sec:motivation}
Dealing with thermal infrared (\textit{IR}) images is important in security and defense tasks like video surveillance or unmanned ground vehicles, and moreover it plays an upcoming role in autonomous driving systems. A trained Convolutional Neural Network (CNN) for this has learned to detect persons, vehicles, buildings, animals or other objects of interest. The properties of the training data that is fed into the model were decisive for the success of detecting objects.  Video cameras should monitor day and night, so our regarded detector should have the ability to classify both types of images data reliable. 

\begin{figure}[t]
	\centering
	\subfloat[][]{\includegraphics[width=0.318\textwidth]{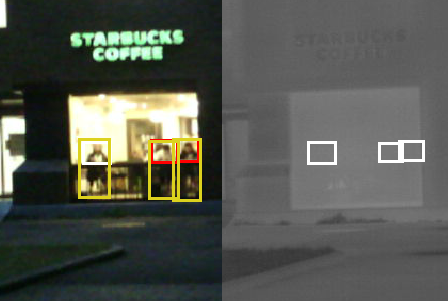}}
	\quad
	\subfloat[][]{\includegraphics[width=0.318\textwidth]{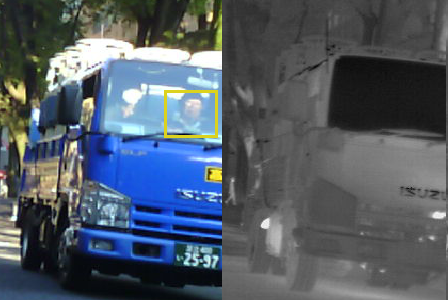}}
	\quad
	\subfloat[][]{\includegraphics[width=0.318\textwidth]{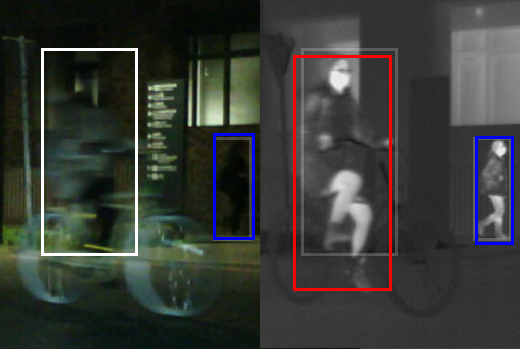}}
	\quad
	\subfloat[][]{\includegraphics[width=0.318\textwidth]{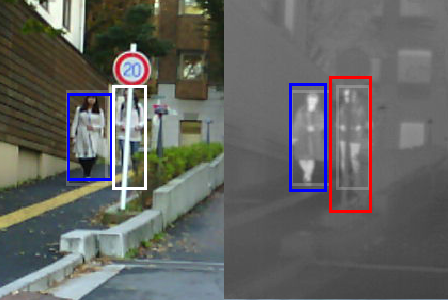}}
	\quad
	\subfloat[][]{\includegraphics[width=0.318\textwidth]{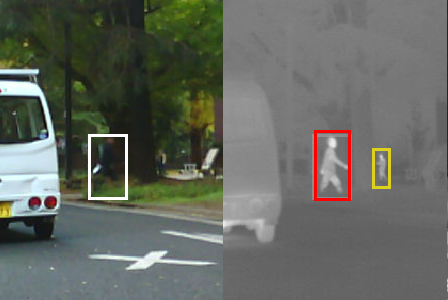}}
	\quad
	\subfloat[][]{\includegraphics[width=0.318\textwidth]{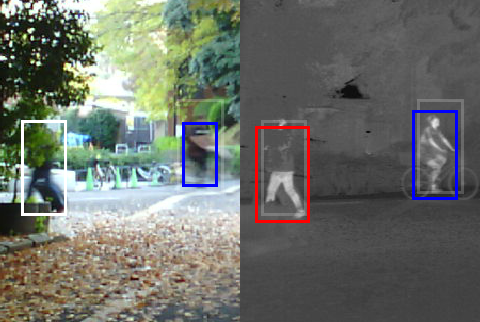}}
	\caption[Examples]
	{\label{fig:examples1}
		Examples for advantages and disadvantages in \textit{IR} or \textit{VIS} detection: Objects behind glass are only visible in \textit{VIS} (a,b). Motion blur in dark situations, where IR works better (c). Diffuse occlusion leads to better results in \textit{IR} (d-f). (White boxes are ground truth, blue boxes detections in \textit{IR} and \textit{VIS}, red boxes detections in only one spectral range, and yellow boxes detections without matching ground truth). All examples from MIL Tokyo detection dataset \cite{takumi2017multispectral}.}
\end{figure}

\begin{figure}[]
	\centering
	\subfloat[][]{\includegraphics[height=0.3\textwidth]{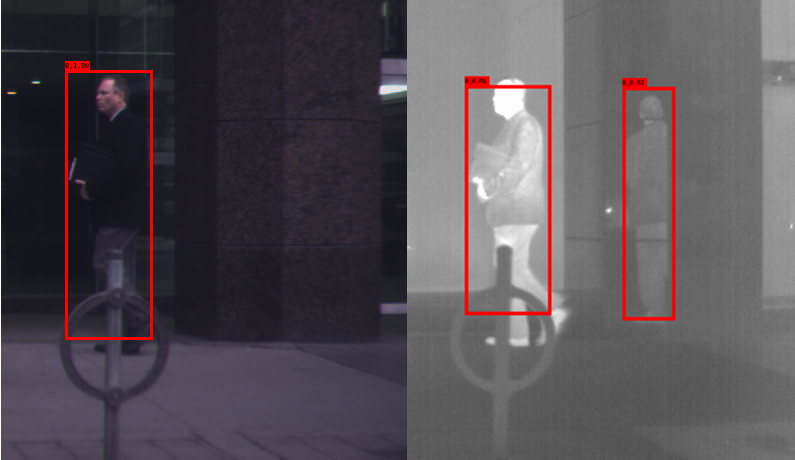}}
	\quad
	\subfloat[][]{\includegraphics[height=0.3\textwidth]{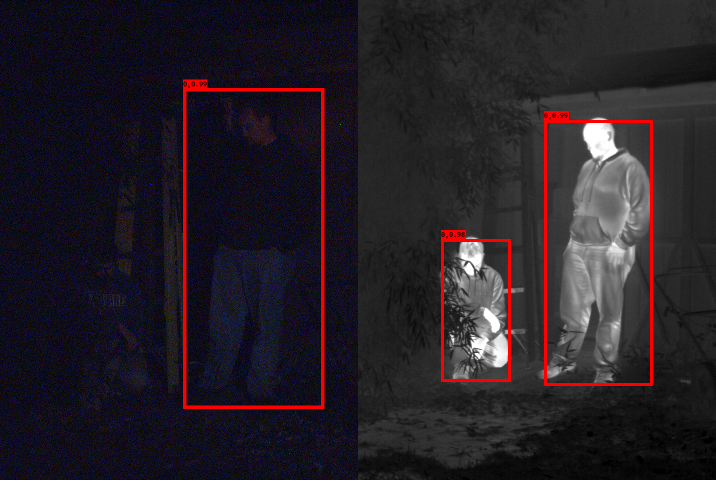}}
	\caption[Examples]
	{\label{fig:examples2}
		More examples for advantages and disadvantages in \textit{IR} or \textit{VIS} detection: \textit{IR} detection is more error-prone due to reflections in glassy surfaces (a). Example from DGB Toronto dataset \cite{morris2007statistics}. \textit{IR} detection works better in darkness (b). Example from CATS dataset \cite{treible2017cats}.}
\end{figure}

\begin{figure}[]
	\centering
	\captionsetup[subfigure]{labelformat=empty}
	\subfloat[][]{\includegraphics[width=0.85\textwidth]{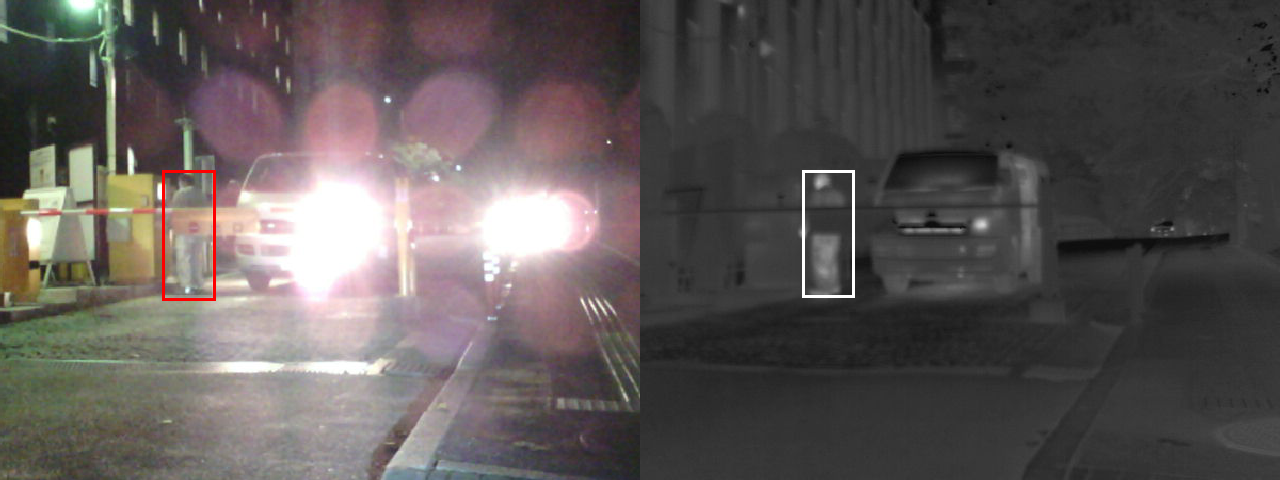}}
	\caption[Example]
	{\label{fig:example}
	  Sometimes the more visible spectrum for humans is not the one with the better detection. Red box: Detection in \textit{VIS}, white box missed ground truth in \textit{IR}. Image from MIL Tokyo detection dataset \cite{takumi2017multispectral}.}
\end{figure}

Depending on the dataset, if a network has been trained with images of the visible spectrum (\textit{VIS}) one could assume that it is good in classifying visible images but not in classifying infrared ones and vice versa. However, indeed, 
\textit{IR} and \textit{VIS} images can enrich each other by the respective advantages or missing disadvantages which are there: Objects behind glass, not visible in \textit{IR}, see Figure \ref{fig:examples1} (a,b), disturbing reflections in \textit{IR} images, see Figure \ref{fig:examples2} (a), diffuse occluded objects in \textit{VIS} images, see Figure \ref{fig:examples1} (d-f), or problems with too little light, see Figures \ref{fig:examples1} (c) and \ref{fig:examples2} (b). It is not always easy for humans to predict what spectral range works well and what does not. For example the person in the \textit{VIS} image in Figure \ref{fig:example} is not easily detected by humans since the image is quite a lot confusing, but better detected from a machine than the one in the corresponding \textit{IR} image.

\begin{figure}[]
	\centering
	\captionsetup[subfigure]{labelformat=empty}
	\subfloat[][]{\includegraphics[width=0.6\textwidth]{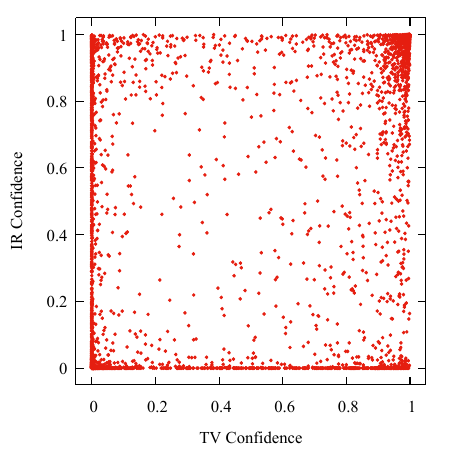}}
	\caption[Confidence values]
	{\label{fig:confidence}
		YOLOv3 Detector on MILtokyo Dataset: Confidence values in \textit{IR} and \textit{VIS} detection. The values of corresponding detections with respect to the ground truth against each other.}
\end{figure}

If one looks at the confidence values of related \textit{IR} and \textit{VIS} detection pairs according to ground truth, it is noticeable that there is still room left for the exploitation of both spectral ranges. The bottom right and the top left corners in Figure \ref{fig:confidence} show detections with no detection or only low confidence detection in the other spectral range. 

Hence, it is not clear, which training data achieve maximum performance in multispectral evaluation. Furthermore, a big problem is that CNNs need a huge set of training data and it is a tedious procedure to annotate it. Especially compatible annotated long-wave-infrared data is rare. That is the reason why we prefer to train our network with images of the visible range and find augmentation techniques to prepare them suitable.

\subsection{Related Work}
In 2008 Zhang and Viola \cite{zhang2008multiple} introduced an infrared person-detector, which was trained on images from the visible spectrum since the overall shape of humans is similar in each spectra. One of early comparative studies on visible and thermal face recognition was presented
by Socolinsky and Selinger \cite{socolinsky2002comparative}. They concluded that "LWIR thermal imagery of human faces is not only a valid biometric, but almost surely a superior one to comparable
visible imagery." In 2013, Kieritz et al. \cite{kieritz2013learning} used a transmodal detector from the visual spectrum for training and applied it on infrared images. 

 Nowadays CNN-based methods became popular because they gain good practical results. They are universal frameworks and need not much development. In 2015, Sarfraz and Stiefelhagen \cite{sarfraz2015deep} published an approach for thermal-to-visible face recognition. Their model attempts to learn a non-linear mapping from visible to thermal spectrum while preserving the identity information.
K{\"o}nig et al. \cite{konig2017fully} presented a novel multispectral Region Proposal Network that is built upon the pre-trained VGG-16 and employed to detect persons from multispectral images that combine visual-optical and infrared image information.
In 2018 Herrmann et al. \cite{herrmann2018cnn} used pre-trained neural networks for \textit{IR} data, which is transformed as close as possible to the \textit{RGB} domain. Their applied methods were stretching, equalization, inversion and combination of stretching and inversion.

\subsection{Contribution}
We take up the method of Herrmann et al. \cite{herrmann2018cnn} and adapt and optimize the augmentation techniques to achieve better results when our test data is a mixture of \textit{VIS} and \textit{IR} data. Contrary to them, we transform the visible data as close as possible to the infrared data and not the other way around. We do not use stretching or equalization, rather we convert the images in different color systems. Our goal is to make easy available train data compatible for a detector, which is specialized on a mixture of visible and thermal infrared images. It is obviously purposeful to make the visible images more looking like infrared ones. Therefor we try several methods that were described and visualized in the following. On the whole, we reach an improvement of about 76\% and 24\% in classification of infrared images and combined images of the visible and thermal infrared spectrum, respectively in comparison to unaugmented visible train data. The performance of only visible test data remains unchanged. To underline the results and to get insights what the CNN learns from the input data, we also compare the filterkernels of the first convolution of the utilized techniques.



\section{Methods}

\subsection{Examples of employed augmentation techniques}
In this work we have tuples of input images $(i,v)\in IR,VIS$, which we describe now. 
Let $T$ be a tensor with entries in $[0,1]$ representing an image with $c$ channels. The pixels of the visible color input image $v$ with height $h$ and width $w$ are scaled to 
\begin{equation}
v\in[R,G,B]\subset T^{h\times w\times c},
\end{equation}
where $c=3,\, v_{ij1}=R,\,v_{ij2}=G,\,v_{ij3}=B\in T^{h\times w}$. Furthermore let
\begin{equation}
i\in IR\subset T^{h\times w}
\end{equation} 
describe thermal infrared images. Now, we want to convert the visible \textit{RGB} images. For a conversion into grayscale one applies a convex combination. 
The model is also known as \textit{Luminance}, and is inspired by the current spectra, which human eyes are differently sensitive to perceiving red, green, and blue:
\begin{equation}
v'=[0.2989R+0.5870G+0.1140B]\in T^{h\times w}.
\label{G_1}
\end{equation}
Another grayscale space is the \textit{Intensity}, mean of the \textit{RGB} channels:
\begin{equation}
v'=\frac{1}{3}[R+G+B]\in T^{h\times w}.
\label{I}
\end{equation}
More simple is \textit{Value}, set pixel-wise
\begin{equation}
v'=\max_{i,j}(v_{ij1},v_{ij2},v_{ij3})\in T^{h\times w}
\end{equation}
by regarding only the maximum entry along the third axis.\\
\textit{Gaussian blur} is also an augmentation technique to make the images more blurring and is the same as convolving each channel of the image with a 2-dimensional Gaussian function. For each channel we calculate pixel-wise:\\
\begin{equation}
v_{ij}'=v_{ij}\ast \frac{1}{2\pi \sigma^2}\exp^{-\frac{\Vert v_{ij}\Vert^2}{2\sigma^2}}.
\label{Gauß}
\end{equation}
To calculate the \textit{inverse} of an image $v\in VIS$ we just set
\begin{equation}
v'=\mathds{1}_{hwc}-v,
\end{equation}
whereby $\mathds{1}_{hwc}$ describes a $h\times w\times c$-tensor with all ones.\\
Intensity can be modified by adding $b\in [-1,1]$ to every entry $v_{ijc}$ in $v$:
\begin{equation}
v'_{ijc}=\mathrm{min}(\mathrm{max}(v_{ijc}+b,0),1).
\end{equation}
To change the \textit{contrast} by factor $k\geq 0$, set channel-wise:
\begin{equation}
v'_{ij}=(v_{ij}-\mathds{1}_{hw}\cdot\mathrm{mean}(v_{ij}))\cdot k +\mathds{1}_{hw}\cdot\mathrm{mean}(v_{ij}).\\
\end{equation}

In this work we apply augmentation techniques $A$ to the input images $v\in VIS$, see Buhrmester et al.\cite{BuhrColor2019}
\begin{equation}
A\colon v\to A(v),
\label{aug}
\end{equation}
$A$ is the successive execution of one or more of the of the described conversions or the identity function. The pre-processed image can now be fed in the classification algorithm, the CNN. It depicts the input images $v\in VIS, i\in IR$ to a prediction:
\begin{equation}
\mathrm{CNN}\colon A(v)\to P',\,\,\,\, \mathrm{CNN}\colon i\to P.
\label{cnn}
\end{equation}
and our goal is to calculate $A$, so that $P\approx P'$.

\section{Experiments}
In this section we describe our data, our deep neural network and how we apply the before presented augmentation techniques to the input images. Furthermore we explain the results of the study.

\subsection{Data}

\begin{figure*}
	\begin{center}
\includegraphics[width=1\linewidth]{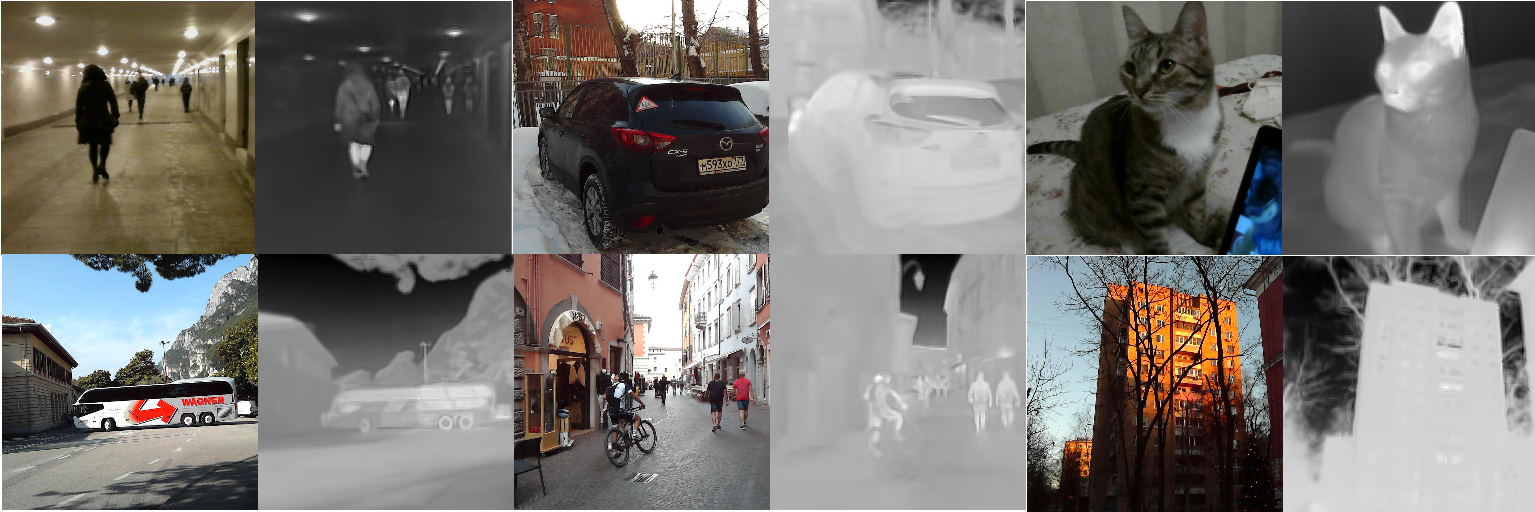}
\end{center}
	\caption{Images of the ThermalWorld Dataset, including humans, cars, cats, bus, humans, building, each in \textit{VIS} and \textit{IR}.}
	\label{thworld}
\end{figure*}

We use the ThermalWorld dataset \cite{kniaz2018thermalgan} for our experiments. It provides 1659 pairs of aligned color and thermal images with ground truth class annotations. Images sizes are $1185\times 889\times 3$ pixels and $640\times 480\times 1$ pixels, respectively. About 21\% are taken for validation. The data is labeled with the 9 classes [car, truck, minibus, bus, building, human, cat, dog, boat], see Figure~\ref{thworld}.

\subsection{Neural Network}

As model we use an established Convolutional Neural Network which performs well in classification tasks. The architecture has two convolutional layers (\textit{conv\,1}, \textit{conv\,2}) with maxpooling (\textit{pool\,1, pool\,2}) and three fully connected layers (\textit{fc\,1}, \textit{fc\,2}, and \textit{softmax}) and further state-of-the-art tools like standardization, data augmentation, normalization, and dropout, see Hinton et al.\cite{hinton2012improving} We have accordingly modified it to enter color, grayscale or infrared images and to visualize the kernels and output of the first layer. The input images are scaled at a size of $128\times128\times3$ and $128\times128\times1$, respectively.

\subsection{Execution}
First we train our neural network with thermal infrared images. After that we train our network with visible \textit{RGB} images. We get a comparative value \textit{P} for each test mode: test images are \textit{IR} or \textit{VIS} or a mix of them (\textit{IR}/\textit{VIS}).
Our goal is to find an augmentation technique $A$ to improve the performance \textit{P'} of the network in the test cases \textit{IR, VIS}, and \textit{IR/VIS}. It is promising to use transformations that transfer the visible images optical as close as possible in thermal infrared images. First we try a conversion into the different grayscale modes Luminance (Lum), Intensity and Value. After that we add Gaussian blurring (blur) with $\sigma=0.5$ on the images. We also invert (inv) them or change the brightness or the contrast. We make experiments with several combinations of those transformations. Finally, we feed these different augmented images into our architecture and train the network in several modes. We compare the accuracies $P'(A)$ for our test images (\textit{IR}, \textit{VIS}, and \textit{IR}/\textit{VIS}) to find out which augmentation $A$ leads closest to $P\approx P'$. In the evaluation the respective same pre-processing is employed to the visible test images \textit{VIS} of course.

\subsection{Results}

\begin{table}[htbp]
\caption{Test accuracy in \% under different augmentation techniques.\\}
\label{xtable1}
\caption{\\}
\centering
\begin{minipage}[c]{0.42 \textwidth}%
\begin{tabular}{|l|c|c|c|}
\hline
train modus / test modus &IR & VIS & IR/VIS\\
\hline\hline
RGB&33&99&66\\
\hline
Luminance&32&99&65.5\\
Lum blur&52&71&61.5\\
Lum inv&54&99&76.5\\
Intensity inv blur&56&99&77.5\\
Lum inv blur contrast&56&99&77.5\\
Lum inv blur &\textbf{58}&\textbf{99}&\textbf{78.5}\\
Lum inv blur brighter&\textbf{58}&\textbf{99}&\textbf{78.5}\\
\hline
\textcolor[rgb]{0.41,0.41,0.41}{IR}&\textbf{\textcolor[rgb]{0.41,0.41,0.41}{76}}&\textbf{\textcolor[rgb]{0.41,0.41,0.41}{51}}&\textbf{\textcolor[rgb]{0.41,0.41,0.41}{63.5}}\\
\hline
\end{tabular}
\end{minipage}\begin{minipage}[c]{0.58 \textwidth}
  \vspace{-\ht\strutbox}\includegraphics[width=10cm]{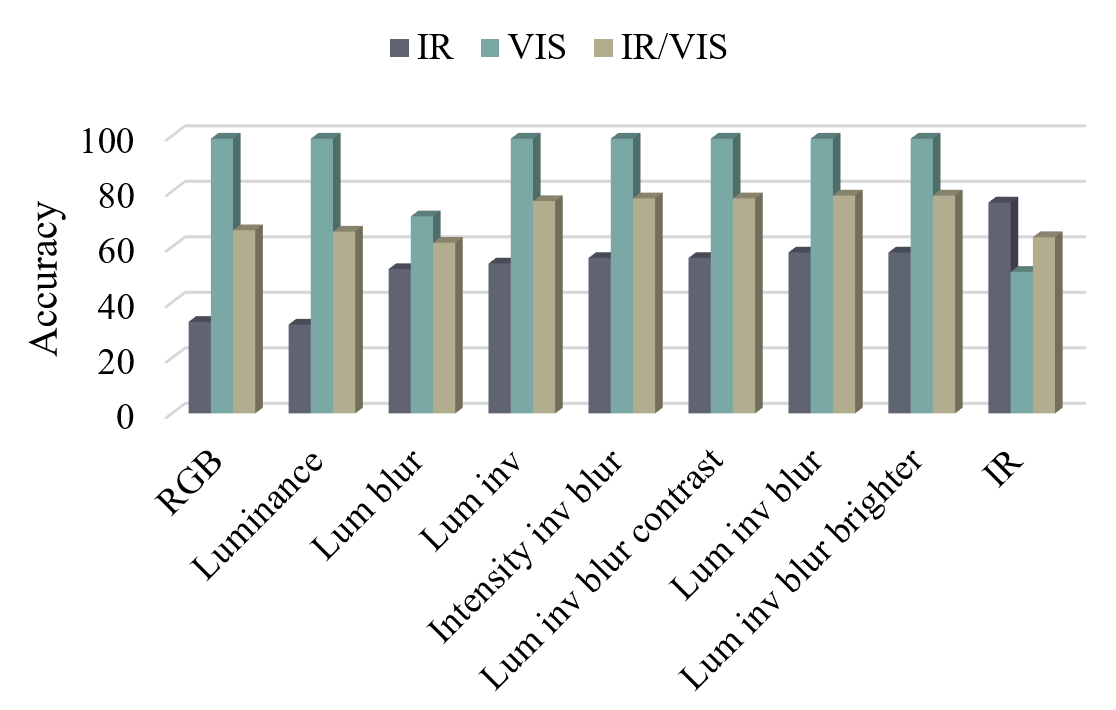}\hfill
\end{minipage}
\def\svgwidth{1\linewidth}
\end{table}

\textbf{Evaluation results.} The important output values are in Table \ref{xtable1}. In the first column are the augmentation techniques for the train and test images from the visible spectrum. In the other columns are the accuracies in different test modes (\textit{IR}, \textit{VIS}, \textit{IR}/\textit{VIS}) shown. \textit{IR/VIS} contains 50\% of each kind of image, that is why the result is the average precision. The graphic illustrates the numbers.

\begin{figure}[t]
	\begin{center}
		\captionsetup[subfigure]{labelformat=empty}
\includegraphics[width=1\linewidth]{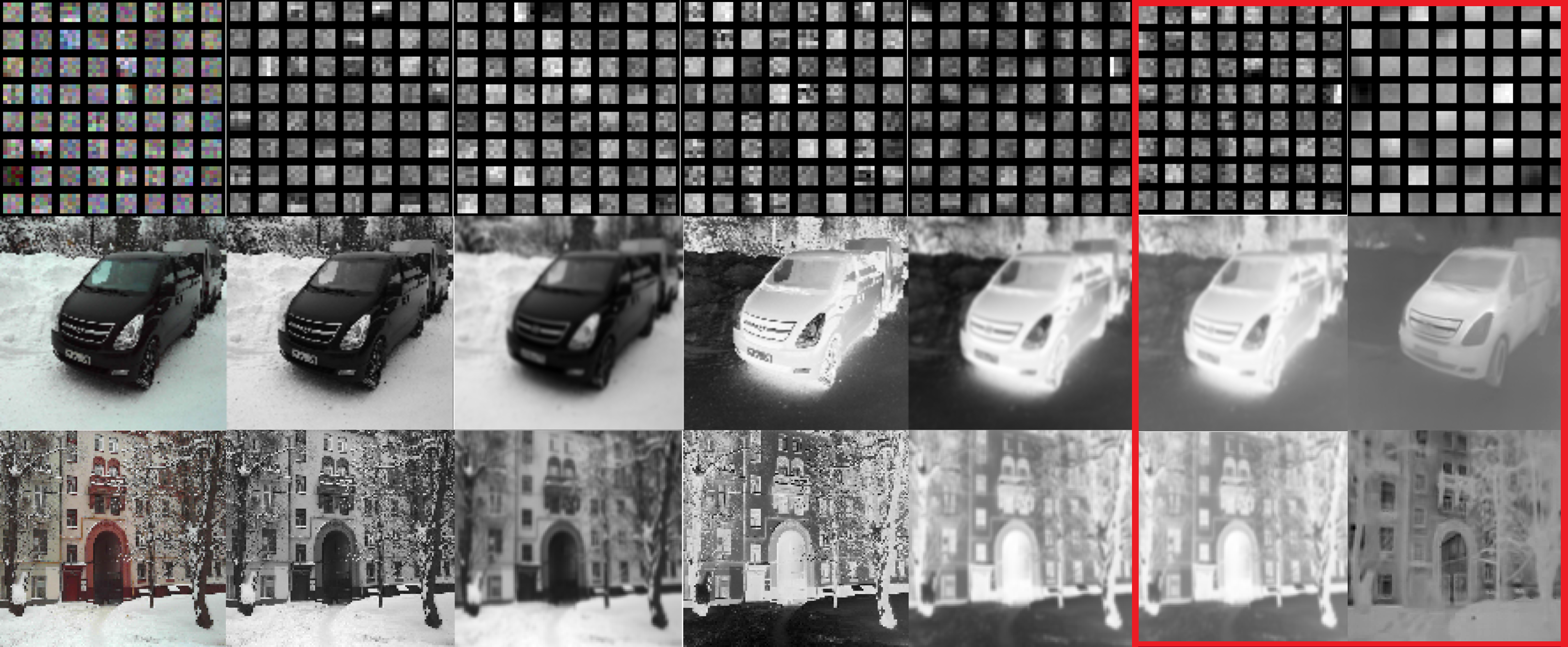}
\end{center}
	\caption{Filterkernels of the first layer, car, and building while training with the following augmentation techniques, from left to right:\\
	(a) none (\textit{RGB}), (b) Luminance, (c) Lum blur, (d) Lum inv, (e) Lum inv blur, (f) Lum inv blur brighter (all \textit{VIS}) and (g) \textit{IR}. \\
	The images in the red box have the greatest similarities.}
	\label{carfilter}
\end{figure}

 Contrary to expectations \textit{RGB} performs better than Luminance even though the infrared spectrum contains no color. Although Gaussian blur with $\sigma=0.5$ significantly improves the detection of \textit{IR} images, the recognition rate of \textit{VIS} images decreases hence blurring is known to degrade shapes. Furthermore, inversion of Luminance images increase the detection rate. This can be gained by additional Gaussian blur. Adjusting contrast seems to be also a good idea, as one compares image \textit{(e)} and \textit{(g)}, Figure \ref{carfilter} but even has a negative effect on the accuracy. Decreasing brightness about 4\% scored no further improvement. As expected Luminance is more suitable than Intensity.

Altogether we reach an improvement of $-1+78.5/63.5\approx 24\%$ when the test data is a mix of \textit{IR} and \textit{VIS} images and the training mode is a combinatino of inversion, Gaussian blurring and Luminance in comparison to infrared training. In comparison to unaugmented visible \textit{RGB} training our approach performs $-1+78.5/66\approx 19\%$ better.

If only \textit{IR} test data is available we achieve a deterioration of $-1+58/76\approx 24\%$  with inverse Luminance and blur train images, but together with the advantage that we can use easily available visible train data. Regarding the use of unaugmented visible train data instead would yield a deterioration of $-1+58/33\approx 76\%$.

To detect objects on images from the visible spectrum, the algorithm performs well with mostly all data augmentation methods applied on the visible data.
\\ 

\textbf{Filter visualization.} 
In Figure \ref{carfilter} the 64 filters in the first convolutional layer are presented under several train modes and six different augmentation techniques.
The 3D filters of training with \textit{RGB} images are colored, the others are 1-channeled because the images are grayscale. One can regard the similarity to the \textit{IR} filters \textit{(g)} is increasingly approaching from \textit{(a)} to \textit{(f)}. This underlines the above results of our study. \\

\section{Conclusion and Future Work}
With our augmentation method we reach an improvement of about 24\% in classification of combined images of the visible and thermal infrared spectrum. The used ThermalWorld dataset includes nine classes, especially persons, vehicles, pets, and buildings and the application is important for video surveillance. Next to the enhancement of the accuracy it is beneficial to use visible train data instead of infrared data which is still available to a limited extend. Our best performing color augmentation method has been the transformation of \textit{RGB} to inverse \textit{Luminance} with added Gaussian blur. Adapting brightness or contrast was contrary to expectation not helpful. Methods like gamma correction or other color systems could be explored further.

\begin{center}
\vfill
{\footnotesize
Copyright 2019 Society of Photo-Optical Instrumentation Engineers. One print or electronic copy may be made for personal use only. Systematic reproduction and distribution, duplication of any material in this paper for a fee or for commercial purposes, or modification of the content of the paper are prohibited.

\url{https://doi.org/10.1117/12.2532537}
}
\end{center}

\bibliography{colorbib} 
\bibliographystyle{spiebib} 

\end{document}